\documentclass[10pt,twocolumn,letterpaper]{article}

\usepackage{cvpr}
\usepackage{times}
\usepackage{epsfig}
\usepackage{graphicx}
\usepackage{amsmath}
\usepackage{amssymb}
\usepackage{blindtext}
\graphicspath{{/}}
\usepackage{ctable}
\DeclareGraphicsExtensions{.pdf,.jpeg,.png,.eps,.jpg,.jpeg}
\usepackage{multirow}
\usepackage{hhline,xcolor}
\usepackage{makecell}
\usepackage{boldline}

\usepackage[breaklinks=true,bookmarks=false]{hyperref}
\hypersetup{linkcolor=red, citecolor=green, colorlinks=true}

\cvprfinalcopy 

\def\cvprPaperID{****} 

\begin{document}

\title{Stereo R-CNN based 3D Object Detection for Autonomous Driving}

\author{Peiliang Li$^1$, Xiaozhi Chen$^2$, and Shaojie Shen$^1$\\
$^1$The Hong Kong University of Science and Technology, $^2$DJI\\
{\tt\small pliap@connect.ust.hk, cxz.thu@gmail.com, eeshaojie@ust.hk}
}

\maketitle

	\begin{abstract}
		We propose a 3D object detection method for autonomous driving by fully exploiting the sparse and dense, semantic and geometry information in stereo imagery. Our method, called Stereo R-CNN, extends Faster R-CNN for stereo inputs to simultaneously detect and associate object in left and right images. We add extra branches after stereo Region Proposal Network (RPN) to predict sparse keypoints, viewpoints, and object dimensions, which are combined with 2D left-right boxes to calculate a coarse\footnote{We use the coarse 3D box to represent one with accurate 2D projection but not necessarily with accurate 3D position.} 3D object bounding box. We then recover the accurate 3D bounding box by a region-based photometric alignment using left and right RoIs. Our method does not require depth input and 3D position supervision, however, outperforms all existing fully supervised image-based methods. Experiments on the challenging KITTI dataset show that our method outperforms the state-of-the-art stereo-based method by around 30\% AP on both 3D detection and 3D localization tasks.
Code has been released at \href{https://github.com/HKUST-Aerial-Robotics/Stereo-RCNN}{https://github.com/HKUST-Aerial-Robotics/Stereo-RCNN}.
	\end{abstract}
	
	\section{Introduction}
	
	3D object detection serves as an essential basis of visual perception, motion prediction, and planning for autonomous driving. Currently, most of the 3D object detection methods \cite{chen2017multi,qi2017frustum,zhou2017voxelnet,ku2017joint,liang2018deep} heavily rely on LiDAR data for providing accurate depth information in autonomous driving scenarios. 
However, LiDAR has the disadvantage of high cost, relatively short perception range ($\sim$100 m), and sparse information (32, 64 lines comparing to $>$720p images). On the other hand, monocular camera provides alternative low-cost solutions\cite{chen2016monocular,mousavian20173d,cvpr18xu} for 3D object detection. The depth information can be predicted by semantic properties in scenes and object size, etc. However, the inferred depth cannot guarantee the accuracy, especially for unseen scenes. To this end, we propose a stereo-vision based 3D object detection method. Comparing with monocular camera, stereo camera provides more precise depth information by left-right photometric alignment. Comparing with LiDAR, stereo camera is low-cost while achieving comparable depth accuracy for objects with non-trivial disparities. The perception range of stereo camera depends on the focal length and the baseline. Therefore, stereo vision has the potential ability to provide larger-range perception by combining different stereo modules with different focal length and baselines.
	
\begin{figure*}
		\begin{center}
			\includegraphics[width=1.91\columnwidth]{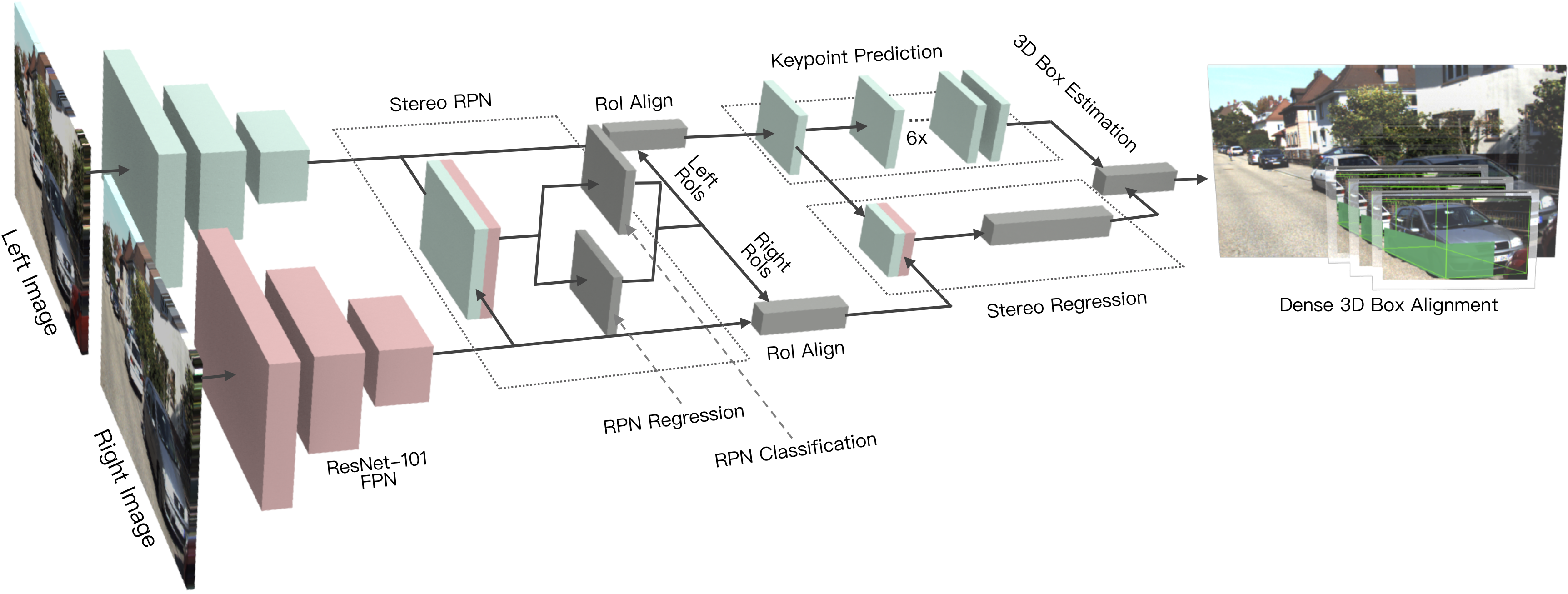}
		\end{center}
		\vspace{-0.2cm}
		\caption{Network architecture of the proposed Stereo R-CNN (Sect.~\ref{sec:rcnn}) which outputs stereo boxes, keypoints, dimensions, and the viewpoint angle, followed by the 3D box estimation (Sect.~\ref{sec:box}) and the dense 3D box alignment module (Sect.~\ref{sec:dense}).}
		\label{fig:system}
	\end{figure*}

	In this work, we study the sparse and dense constraints for 3D objects by fully exploiting the semantic and geometry information in stereo imagery and propose an accurate Stereo R-CNN based 3D object detection method.
	Our method simultaneously detects and associates objects for left and right images using the proposed Stereo R-CNN. 
	The network architecture can be overviewed in Fig.\ref{fig:system}, which can be divided into three main parts. The first one is a Stereo RPN module (Sect.~\ref{sec:rpn}) which outputs corresponding left and right RoI proposals. 
	After applying RoIAlign \cite{he2017mask} on left and right feature maps respectively, we concatenate left-right RoI features to classify object categories and regress accurate 2D stereo boxes, viewpoint, and dimensions in the stereo regression (Sect.~\ref{sec:stereobox}) branch.
	A keypoint (Sect.~\ref{sec:keypoint}) branch is employed to predict object keypoints using only left RoI feature. 
	These outputs form the sparse constraints (2D boxes, keypoints) for the 3D box estimation (Sect.~\ref{sec:box}), where we formulate the projection relations between 3D box corners with 2D left-right boxes and keypoints. 
	
	The crucial component that ensures our 3D localization performance is the dense 3D box alignment (Sect.~\ref{sec:dense}). We consider 3D object localization as a learning-aided geometry problem rather than an end-to-end regression problem. Instead of directly using the depth input \cite{3dopJournal,cvpr18xu} which does not explicitly utilize the object property, we treat the object RoI as an entirety rather than independent pixels. For regular-shaped objects, the depth relation between each pixel and the 3D center can be inferred given the coarse 3D bounding box. We warp dense pixels in the left RoI to the right image according to their depth relations with the 3D object center to find the best center depth that minimizes the entire photometric error. The entire object RoI thereby forms the dense constraint for 3D object depth estimation. 
	The 3D box is further rectified using 3D box estimator (Sect.~\ref{sec:box}) according to the aligned depth and 2D measurements.
	
	We summarize our main contributions as follows:
	\begin{itemize}
		\item A Stereo R-CNN approach which simultaneously detects and associates object in stereo images.
		\item A 3D box estimator which exploits the keypoint and stereo boxes constraints.
		\item A dense region-based photometric alignment method that ensures our 3D object localization accuracy.
		\item Evaluation on the KITTI dataset shows we outperform all state-of-the-art image-based methods and are even comparable with a LiDAR-based method \cite{li2016vehicle}. 
	\end{itemize}
	
	\section{Related Work}
	We briefly review recent works of 3D object detection based on the LiDAR data, monocular image and stereo images respectively.
\vspace{-0.2cm}	
	{\setlength{\parindent}{0cm}
		\subparagraph*{LiDAR-based 3D Object Detection.}Most of the state-of-the-art 3D object detection methods rely on LiDAR to provide accurate 3D information, while process raw LiDAR input in different representations. \cite{chen2017multi,li2016vehicle,yang2018pixor,liang2018deep, ku2017joint} project the point cloud into 2D bird's eye view or front view representations and feed them into the structured convolution network, where \cite{chen2017multi,liang2018deep, ku2017joint} exploit fusing multiple LiDAR representations with the RGB image to obtain more dense information. \cite{engelcke2017vote3deep,wang2015voting,li20173d,luo2018fast,zhou2017voxelnet} utilize structured voxel grid representation to quantize the raw point cloud data, then use either 2D or 3D CNN to detect 3D object, while\cite{luo2018fast} takes multiple frames as input and generates 3D detection, tracking and motion forecasting simultaneously. Additionally, instead of quantizing the point cloud, \cite{qi2017frustum} directly takes raw point cloud as input to localize 3D object based on the frustum region reasoned from 2D detection and PointNet \cite{qi2017pointnet}.
	}
\vspace{-0.6cm}
	{\setlength{\parindent}{0cm}
		\subparagraph*{Monocular-based 3D Object Detection.}\cite{chen2016monocular} focuses on 3D object proposals generation using ground-plane assumption, shape prior, contextual feature and instance segmentation from the monocular image. \cite{mousavian20173d} proposes to estimate 3D box using the geometry relations between 2D box edges and 3D box corners.  
		\cite{zeeshan2014cars,chabot2017deep, murthy2017reconstructing} explicitly utilize sparse information by predicting series of keypoints of regular-shape vehicles. The 3D object pose can be constrained by wireframe template fitting. 
		\cite{cvpr18xu} proposes an end-to-end multi-level fusion approach to detect 3D object by concatenating the RGB image and the monocular-generated depth map.
		Recently an inverse-graphics framework \cite{kundu20183d} is proposed to predict both the 3D object pose and instance-level segmentation by graphic rendering and comparing. However, monocular-based methods unavoidably suffer from the lack of accurate depth information.
	}
\vspace{-0.3cm}
	{\setlength{\parindent}{0cm}
		\subparagraph*{Stereo-based 3D Object Detection.}There are surprisingly only a few works exploit utilizing stereo vision for 3D object detection. 3DOP \cite{3dopJournal} focuses on generating 3D proposals by encoding object size prior, ground-plane prior and depth information (e.g., free space, point cloud density) into an energy function. 3D Proposals are then used to regress the object pose and 2D boxes using the R-CNN approach. \cite{li2018semantic} extends the Structure from Motion (SfM) approach to the dynamic object case and continuously track the 3D object and ego-camera pose by fusing both spatial and temporal information. However, none of the above approaches takes advantage of dense object constraints in raw stereo images.
	}
	
	\section{Stereo R-CNN Network}
	\label{sec:rcnn}
	In this section, we describe the Stereo R-CNN network architecture. Compared with the single frame detector such as Faster R-CNN \cite{ren2015faster}, Stereo R-CNN can simultaneously detect and associate 2D bounding boxes for left and right images with minor modifications. We use weight-share ResNet-101 \cite{he2016deep} and FPN \cite{lin2017feature} as our backbone network to extract consistent features on left and right images. Benefit from our training target design Fig.~\ref{fig:target}, there is no additional computation for data association.
	
	\subsection{Stereo RPN}
	\label{sec:rpn}
	Region Proposal Network (RPN) \cite{ren2015faster} is a sliding-window based foreground detector. After feature extraction, a $3 \times 3$ convolution layer is utilized to reduce channel, followed by two sibling fully-connected
	layer to classify objectness and regress box offsets for each input location which is anchored with pre-define multiple-scale boxes. Similar with FPN \cite{lin2017feature}, we modify origin RPN for pyramid features by evaluating  anchors on multiple-scale feature maps. The difference is we concatenate left and right feature maps at each scale, then we feed the concatenated features into the stereo RPN network. 
	\begin{figure}
	\setlength{\belowcaptionskip}{-0.3cm}
		\begin{center}
			\includegraphics[width=1.0\columnwidth]{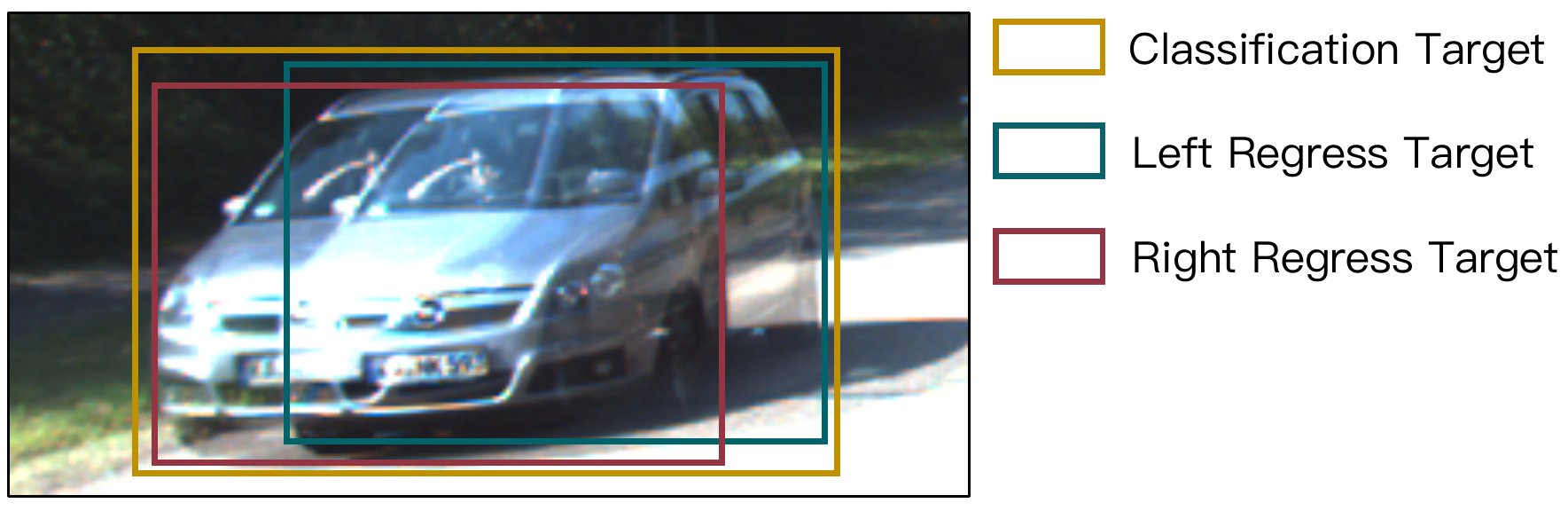}
		\end{center}
		\caption{Different targets assignment for RPN classification and regression.}
		\label{fig:target}
	\end{figure}
	
	The key design enables our simultaneous object detection and association is the different ground-truth (GT) box assignment for objectness classifier and stereo box regressor. As illustrated in Fig.~\ref{fig:target}, we assign the union of left and right GT boxes (referred as union GT box) as the target for objectness classification. An anchor is assigned a positive label if its Intersection-over-Union (IoU) ratio with one of union GT boxes is above 0.7, and a negative label if its IoU with any of union boxes is below 0.3. Benefit from this design, the positive anchors tend to contain both left and right object regions. We calculate offsets of positive anchors respecting to the left and right GT boxes contained in the target union GT box, then assign offsets to the left and right regression respectively. There are six regressing terms for the stereo regressor: $[\Delta u, \Delta w, \Delta u', \Delta w', \Delta v, \Delta h]$, where we use $u,v$ to denote the horizontal and vertical coordinates of the 2D box center in image space, $w, h$ for width and height of the box, and the superscript $(\cdot)'$ for corresponding terms in the right image. Note that we use same $v, h$ offsets $\Delta v, \Delta h$ for the left and right boxes because we use rectified stereo images. Therefore we have six output channels for stereo RPN regressor instead of four in the origin RPN implementation. Since the left and right proposals are generated from the same anchor and share the objectness score, they can be associated naturally one by one. We utilize Non-Maximum Suppression (NMS) on left and right RoIs separately to reduce redundancy, then choose top 2000 candidates from entries which are kept in both left and right NMS for training. For testing, we choose only top 300 candidates.
	
	\begin{figure}
		\begin{center}
			\includegraphics[width=0.8\columnwidth]{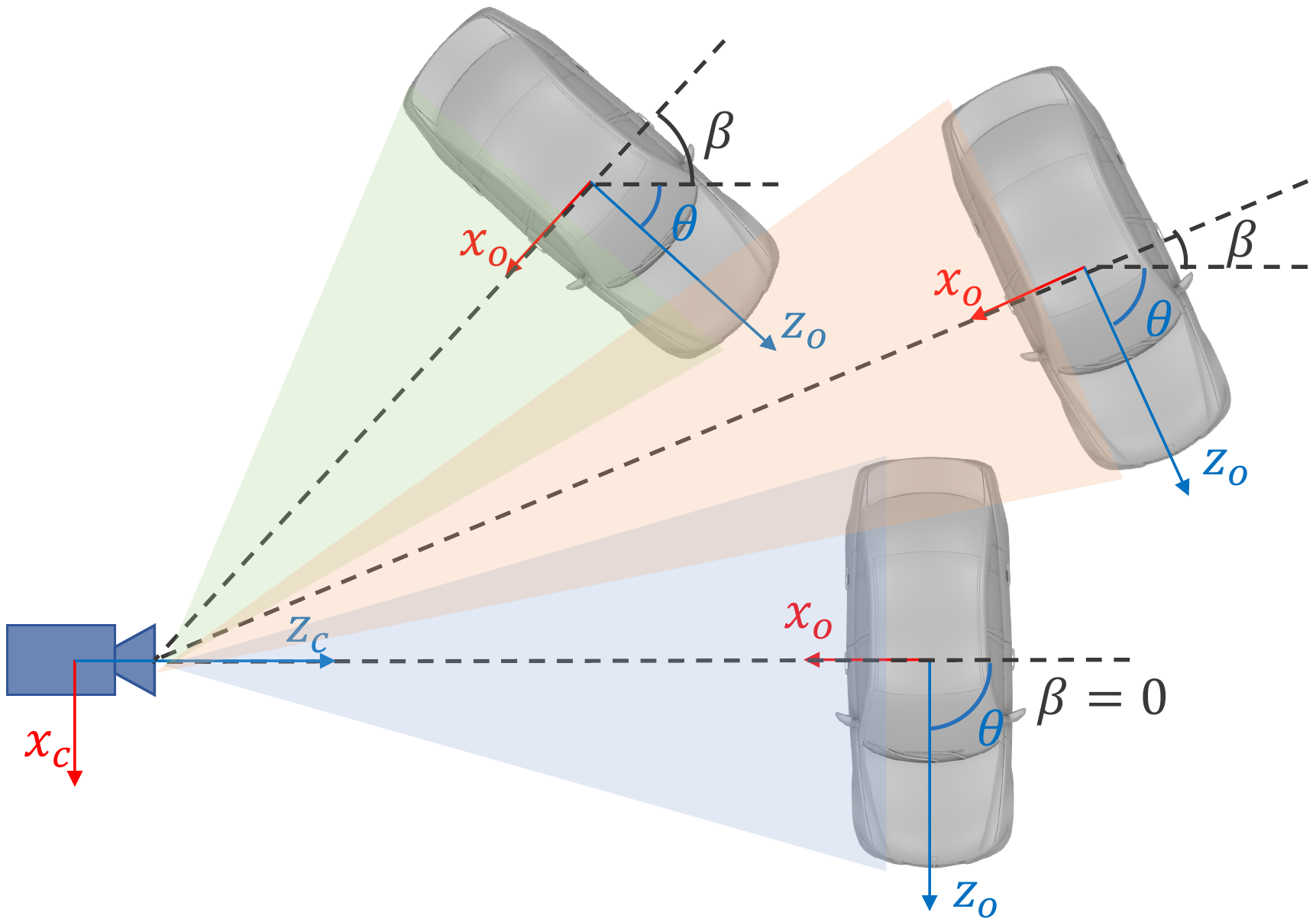}
		\end{center}
		\caption{Relations between object orientation $\theta$, azimuth $\beta$ and viewpoint $\theta+\beta$. Only same viewpoints lead to same projections.}
		\label{fig:angle}
	\end{figure}
	
	\subsection{Stereo R-CNN}
	{\setlength{\parindent}{0cm}
		\subparagraph*{Stereo Regression.}
		\label{sec:stereobox}After stereo RPN, we have corresponding left-right proposal pairs. We apply RoI Align \cite{he2017mask} on the left and right feature maps respectively at appropriate pyramid level. The left and right RoI features are concatenated and fed into two sequential fully-connected layers (each followed by a ReLU layer) to extract semantic information. We use four sub-branches to predict object class, stereo bounding boxes, dimension, and viewpoint angle respectively. The box regression terms are same as defined in Sect.~\ref{sec:rpn}. Note that the viewpoint angle is not equal to the object orientation which is unobservable from cropped image RoI. An example is illustrated in Fig.~\ref{fig:angle}, where we use $\theta$ to denote the vehicle orientation respecting to the camera frame, and $\beta$ to denote the object azimuth respecting to the camera center. Three vehicles have different orientations, however, the projection of them are exactly the same on cropped RoI images. We therefore regress the viewpoint angle $\alpha$ defined as: $\alpha=\theta+\beta$. To avoid the discontinuity, the training targets are $[\sin \alpha, \cos \alpha]$ pair instead of the raw angle value. With stereo boxes and object dimension, the depth information can be recovered intuitively, and the vehicle orientation can also be solved by decoupling the relations between the viewpoint angle with the 3D position.
	}
	
	When sampling the RoIs, we consider a left-right RoI pair as foreground if the maximum IoU between the left RoI with left GT boxes is higher than 0.5, meanwhile the IoU between right RoI with the corresponding right GT box is also higher than 0.5. A left-right RoI pair is considered as background if the maximum IoU for either the left RoI or the right RoI lies in the [0.1, 0.5) interval. For foreground RoI pairs, we assign regression targets by calculating offsets between the left RoI with the left GT box, and offsets between the right RoI with the corresponding right GT box. We still use the same $\Delta v, \Delta h$ for left and right RoIs. For dimension prediction, we simply regress the offset between the ground-truth dimension with a pre-set dimension prior.
	
	\begin{figure}
		\begin{center}
			\includegraphics[width=1.0\columnwidth]{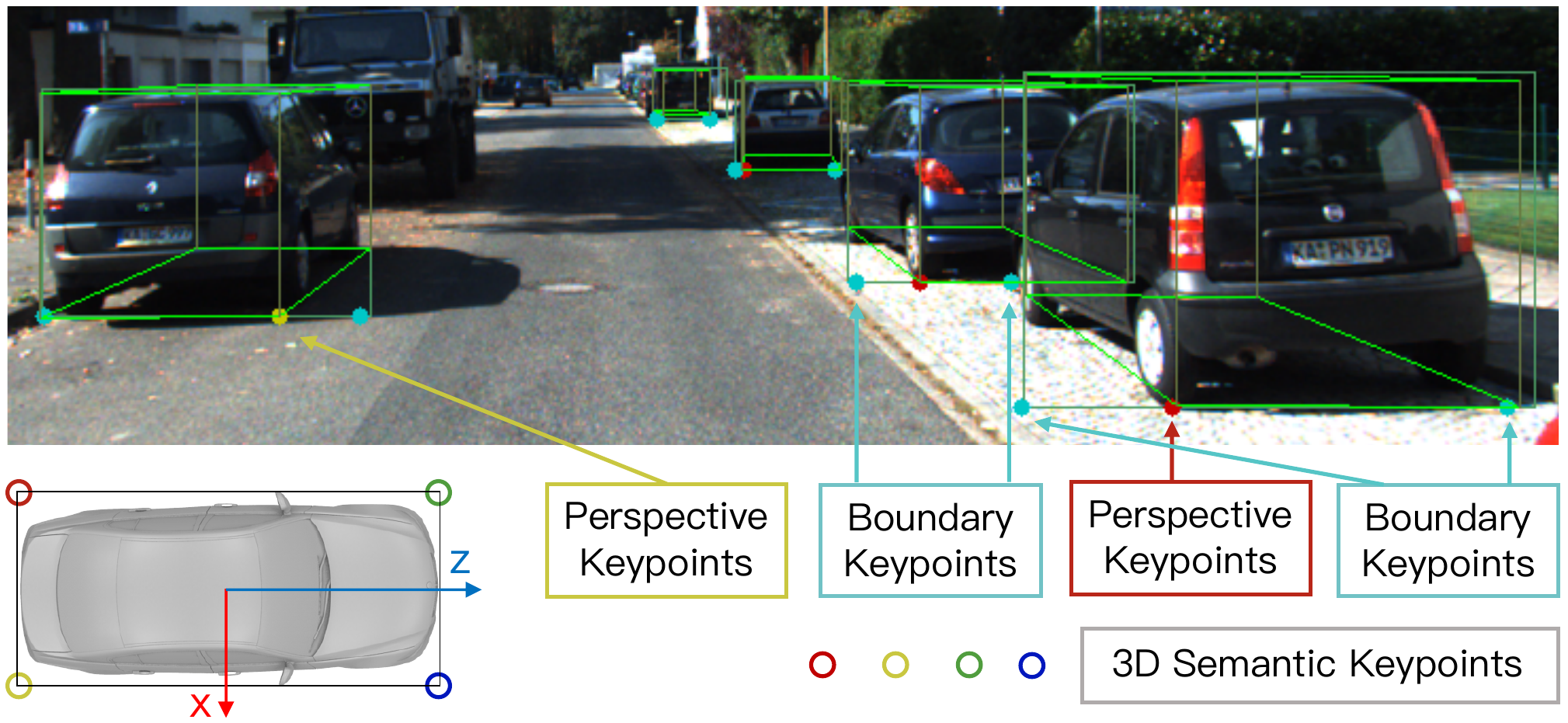}
		\end{center}
		\caption{Illustration of 3D semantic keypoints, the 2D perspective keypoint, and boundary keypoints.}
		\label{fig:keypoint}
	\end{figure}
	{\setlength{\parindent}{0cm}
		\subparagraph*{Keypoint Prediction.}
		\label{sec:keypoint}
		Besides stereo boxes and viewpoint angle, we notice that the 3D box corner which projected in the box middle can provide more rigorous constraints to the 3D box estimation. As Fig.~\ref{fig:keypoint} presents, we define four 3D semantic keypoints which indicate four corners at the bottom of the 3D bounding box. There is only one 3D semantic keypoint can be visibly projected to the box middle (instead of left or right edges). We define the projection of this semantic keypoint as perspective keypoint. We show how the perspective keypoint contributes to the 3D box estimation in Sect.~\ref{sec:box} and Table.~\ref{tab:orien}. We also predict two boundary keypoints which serve as simple alternatives to instance mask for regular-shaped objects. Only the region between two boundary keypoints belongs to the current object and will be used for the further dense alignment (See Sect.~\ref{sec:dense}). 
	}
	
	We predict the keypoint as proposed in Mask R-CNN \cite{he2017mask}. Only the left feature map is used for keypoint prediction. We feed the $14 \times 14$ RoI aligned feature maps to six sequential 256-d $3 \times 3$ convolution layers as illustrated in Fig.~\ref{fig:system}, each followed by a ReLU layer. A $2 \times 2$ deconvolution layer is used to upsample the output scale to $28 \times 28$. We notice that only u coordinate of the keypoints provide additional information besides the 2D box.
	To relax the task, we sum the height channel in the $6 \times 28 \times 28$ output to produce $6 \times 28 $ prediction. As a result, each column in the RoI feature will be aggregated and contribute to the keypoint prediction. The first four channels represent the probability that each of four semantic keypoints is projected to the corresponding u location. The other two channels represent the probability of each u lies in the left and right boundary respectively.
	Note that only one of four 3D keypoints can be visibly projected to the 2D box middle, thereby softmax is applied to the  $4 \times 28 $ output to encourage that one exclusive semantic keypoint is projected to a single location. This strategy avoids the probable confusion of perspective keypoint type (corresponding to which of semantic keypoints). For the left and right boundary keypoints, we apply softmax on the $1 \times 28$ outputs respectively.
	
	During training, we minimize the cross-entropy loss over $4 \times 28$ softmax output for perspective keypoint prediction. Only a single location in the $4 \times 28$ output is labeled as perspective keypoint target. We omit the case where no 3D semantic keypoint is visibly projected in the box middle (e.g., truncation and orthographic projection cases). For boundary keypoints, we minimize the cross-entropy loss over two $1 \times 28$ softmax outputs independently. Each foreground RoI will be assigned the left and right boundary keypoints according to the occlusion relations between GT boxes.
	\section{3D Box Estimation}
	\label{sec:box}
\begin{figure}
		\begin{center}
			\includegraphics[width=1.0\columnwidth]{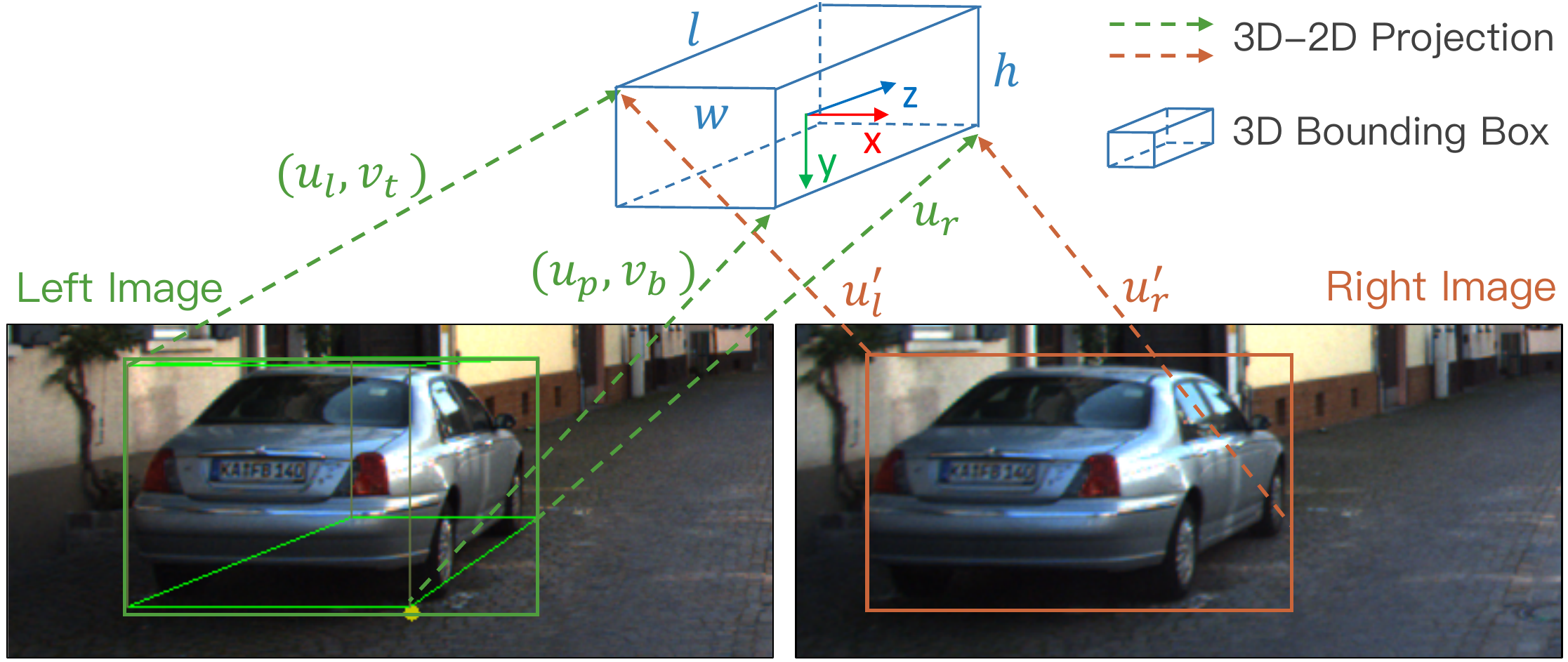}
		\end{center}
		\caption{Sparse constraints for the 3D box estimation (Sect.~\ref{sec:box}).}
		\label{fig:box}
	\end{figure}

	In this section, we solve a coarse 3D bounding box by utilizing the sparse keypoint and 2D box information. States of the 3D bounding box can be represented by $\mathbf{x} = \{x, y, z, {\theta}\}$, which denotes the 3D center position and horizontal orientation respectively. Given the left-right 2D boxes, perspective keypoint, and regressed dimensions, the 3D box can be solved by minimize the reprojection error of 2D boxes and the keypoint. As detailed in Fig.~\ref{fig:box}, we extract seven measurements from stereo boxes and perspective keypoints: $\mathbf{z} = \{u_{l}, v_{t}, u_{r}, v_{b}, u'_{l}, u'_{r}, u_{p}\}$, which represent left, top, right, bottom edges of the left 2D box, left, right edges of the right 2D box, and the u coordinate of the perspective keypoint. Each measurement is normalized by camera intrinsic for simplifying representation. Given the perspective keypoint, the correspondences between 3D box corners and 2D box edges can be inferred (See dotted lines in Fig.~\ref{fig:box}). Inspired from \cite{li2018semantic}, we formulate the 3D-2D relations by projection transformations. In such a viewpoint in Fig.~\ref{fig:box}:
	\begin{equation}
	\label{eq:prior}
	\renewcommand*{\arraystretch}{1.5}
	\begin{array}{lr}
	v_{t} = (y-\frac{h}{2})/(z-\frac{w}{2}sin\theta-\frac{l}{2}cos\theta),\\
	u_{l} = (x-\frac{w}{2}cos\theta-\frac{l}{2}sin\theta)/(z+\frac{w}{2}sin\theta-\frac{l}{2}cos\theta),\\
	u_{p} = (x+\frac{w}{2}cos\theta-\frac{l}{2}sin\theta)/(z-\frac{w}{2}sin\theta-\frac{l}{2}cos\theta),\\
	\qquad \qquad \qquad \qquad \qquad \ldots \\
	u'_{r} = (x-b+\frac{w}{2}cos\theta+\frac{l}{2}sin\theta)/(z-\frac{w}{2}sin\theta+\frac{l}{2}cos\theta).\\
	\end{array}
	\end{equation}
	We use $b$ to denote the baseline length of the stereo camera, and $w, h, l$ for regressed dimensions. There are total seven equations corresponding to seven measurements, where the sign of $\{\frac{w}{2},\frac{l}{2}\}$ should be changed appropriately based on the corresponding 3D box corner. Truncated edges are dropped on above seven equations. These multivariate equations are solved via Gauss-Newton method. Different from \cite{li2018semantic} using single 2D box and size prior to solve the 3D position and orientation, we recover the 3D depth information more robustly by jointly utilizing the stereo boxes and regressed dimensions. In some cases where less than two side-surfaces can be completely observed and no perspective keypoint $u_p$ (e.g., truncation, orthographic projection), the orientation and dimensions are unobservable from pure geometry constraints. We use the viewpoint angle $\alpha$ to compensate the unobservable states (See Fig.~\ref{fig:angle} for the illustration):
	\begin{equation}
	\label{eq:alpha}
	\renewcommand*{\arraystretch}{1.5}
	\begin{array}{lr}
	\alpha = \theta + \arctan (-\frac{x}{z}).
	\end{array}
	\end{equation}
	
	Solved from 2D boxes and the perspective keypoint, the coarse 3D box has accurate projection and is well aligned with the image, which enables our further dense alignment.
	\section{Dense 3D Box Alignment}
	\label{sec:dense}
	The left and right bounding boxes provide object-level disparity information such that we can solve the 3D bounding box roughly. However, the stereo boxes are regressed by aggregating the high-level information in a $7\times7$ RoI feature maps. The pixel-level information (e.g., corners, edges) contained in original image is lost due to multiple convolution filters. To achieve sub-pixel matching accuracy, we retrieve the raw image to exploit the pixel-level high-resolution information. Note that our task is different with pixel-wise disparity estimation problem where the result might encounter either discontinuity at ill-posed regions (SGM \cite{hirschmuller2008stereo}), or oversmooth at edge areas (CNN based methods \cite{zbontar2016stereo,kendall2017end,chang2018pyramid}). We only solve the disparity of the 3D bounding box center while using the dense object patch, i.e., we use plenty of pixel measurements to solve one single variable.
	
	Treating the object as a regular-shaped cube, we know the depth relation between each pixel with the center of 3D bounding box solved from Sect.~\ref{sec:box}. To exclude the pixel belonging to the background or other objects, we define a valid RoI as the region is between the left-right boundary keypoints and lies in the bottom halves of the 3D box since the bottom halves of vehicles fits the 3D box more tightly (See Fig.~\ref{fig:system}). For a pixel located at the normalized coordinate $(u_i,v_i)$  in the valid RoI of the left image, the photometric error can be defined as:
	\begin{equation}
	\label{eq:prior}
	\renewcommand*{\arraystretch}{1.5}
	\begin{array}{lr}
	\mathbf{e}_i = \left\| I_l(u_i,v_i)- I_r(u_i-\frac{b}{z+\Delta z_i},v_i)\right \|,
	\end{array}
	\end{equation}
	where we use $I_l, I_r$ to denote the 3-channels RGB vector of left and right image respectively; $\Delta z_i = z_i - z$ the depth differences of pixel $i$ with the 3D box center; and $b$ the baseline length. $z$ is the only objective variable we want to solve. We use bilinear interpolation to get sub-pixel value on the right image. The total matching cost is defined as the Sum of Squared Difference (SSD) over all pixels in the valid RoI: 
	\begin{equation}
	\label{eq:prior}
	\renewcommand*{\arraystretch}{1.5}
	\begin{array}{lr}
	\mathbf{E} = \sum^{N}_{i=0} \mathbf{e}_i.
	\end{array}
	\end{equation}
		\begin{table*}
		\begin{center}
			\renewcommand{\arraystretch}{1.2}
			\resizebox{0.81\textwidth}{!}{%
				\begin{tabular}{l | ccc | ccc |ccc| ccc}
					\, & \,&  \,& \, & \multicolumn{9}{c}{AP$\rm _{2d}$ (IoU=0.7)} \\
					\cline{5-13}
					\,& \multicolumn{3}{c|}{AR (300 Proposals)}& \multicolumn{3}{c|}{Left} & \multicolumn{3}{c|}{Right} & \multicolumn{3}{c}{Stereo} \\
					\cline{2-13}
					Method & Left & Right & Stereo & \,Easy\, & Mode & \,Hard\, & \,Easy\, & Mode & \,Hard\, &  \,Easy\, & Mode & \,Hard \, \\
					\Xhline{1pt} 
					Faster R-CNN\cite{ren2015faster} & 86.08 & - & - & 98.57 & \textbf{89.01} & \textbf{71.54} & - & - & - & - & - & -\, \\
					Stereo R-CNN$\rm _{mean}$ & 85.50 & 85.56 & 74.60 & 90.58 & 88.42 & 71.24 & 90.59 & 88.47 & 71.28 & 90.53 & 88.24 & 71.12\, \\
					Stereo R-CNN$\rm _{concat}$ & \textbf{86.20} & \textbf{86.27} & \textbf{75.51} & \textbf{98.73} & 88.48 & 71.26 & \textbf{98.71} & \textbf{88.50} & \textbf{71.28} & \textbf{98.53} & \textbf{88.27} & \textbf{71.14}\, \\
				\end{tabular}
			}		
		\end{center}
		\caption{Average recall (AR) (in \%) of RPN and Average precision (AP) (in \%) of 2D detection, evaluated on the KITTI \textit{validation set}. We compare two fusion methods for Stereo-RCNN with Faster R-CNN using the same backbone network, hyper-parameters, and augmentation. The Average Recall is evaluated on the \textit{moderate set}.}
		\label{table:2d}
	\end{table*}	
	\begin{table*}
		\begin{center}
			\renewcommand{\arraystretch}{1.2}
			\resizebox{0.86\textwidth}{!}{%
				\begin{tabular}{l|c|ccc|ccc|ccc|ccc}
					\, & \, &
					\multicolumn{3}{c}{AP$\rm _{bv}$ (IoU=0.5)} & \multicolumn{3}{c|}{AP$\rm _{bv}$ (IoU=0.7)} & \multicolumn{3}{c}{AP$\rm _{3d}$ (IoU=0.5)} & \multicolumn{3}{c}{AP$\rm _{3d}$ (IoU=0.7)} \\
					\cline{3-14}
					Method & Sensor & \,Easy\, & Mode & \,Hard\, &  \,Easy\, & Mode & \,Hard\, & \,Easy\, & Mode & \,Hard\, &  \,Easy\, & Mode & \,Hard \, \\
					\Xhline{1pt}  
					Mono3D\cite{chen2016monocular}  & Mono & 30.50 & 22.39 & 19.16 & 5.22 & 5.19 & 4.13 & 25.19 & 18.20 & 15.52 & 2.53 & 2.31 & 2.31\, \\
					Deep3DBox\cite{mousavian20173d}  & Mono & 30.02 & 23.77 & 18.83 & 9.99 & 7.71 & 5.30 & 27.04 & 20.55 & 15.88 & 5.85 & 4.10 & 3.84\, \\
					Multi-Fusion\cite{cvpr18xu}  & Mono & 55.02 & 36.73 & 31.27 & 22.03 & 13.63 & 11.60 & 47.88 & 29.48 & 26.44 & 10.53 & 5.69 & 5.39\, \\
					
					\hline
					VeloFCN\cite{li2016vehicle} & LiDAR & 79.68 & 63.82 & \textbf{62.80} & 40.14 & 32.08 & 30.47 & 67.92 & 57.57 & {52.56} & 15.20 & 13.66 & 15.98 \,\\
					\hline
					Multi-Fusion\cite{cvpr18xu} & Stereo & - & 53.56 & - & - & 19.54 & - & - & 47.42 & - & - & 9.80 & - \,\\
					3DOP\cite{3dopJournal} & Stereo & 55.04 & 41.25 & 34.55 & 12.63 & 9.49 & 7.59 & 46.04 & 34.63 & 30.09 & 6.55 & 5.07 & 4.10 \,\\
					Ours & Stereo & \textbf{87.13} & \textbf{74.11} & 58.93 & \textbf{68.50} & \textbf{48.30} & \textbf{41.47} & \textbf{85.84} & \textbf{66.28} & \textbf{57.24} & \textbf{54.11} & \textbf{36.69} & \textbf{31.07}\, \\
				\end{tabular}
			}		
		\end{center}
		\caption{Average precision of bird's eye view (AP$_{\rm bv}$) and 3D boxes (AP$_{\rm 3d}$) comparison, evaluated on the KITTI \textit{validation set}.}
		\label{table:SOTA}
	\end{table*}The center depth $z$ can be solved by minimizing the total matching cost $\mathbf{E}$, we can enumerate the depth efficiently to find a depth that minimizes the cost. We initially enumerate 50 depth values around the initial value with 0.5-meter interval to get a rough depth and finally enumerate 20 depth values around the rough depth with 0.05-meter interval to get the accurately aligned depth. Afterwards, we rectify the entire 3D box using our 3D box estimator by fixing the aligned depth (See Table.~\ref{table:alignment}). Consider the object RoI as a geometric-constraint entirety, our dense alignment method naturally avoids the discontinuity and ill-posed problems in stereo depth estimation, and is robust to intensity variations and brightness dominant since each pixel in the valid RoI will contribute to the object depth estimation. Note that this method is efficient and can be a light-weight plug-in module for any image-based 3D detection to achieve depth rectifying. Although the 3D object does not fit the 3D cube rigorously, relative depth errors caused by the shape variation are much more trivial than the global depth. Therefore our geometry-constraint dense alignment provides accurate depth estimation of object center.
	\section{Implementation Details}
	
	{\setlength{\parindent}{0cm}
		\subparagraph*{Network.} 
		As implemented in \cite{ren2015faster}, we use five scale anchors of \{32, 64, 128, 126, 512\} with three ratios \{0.5, 1, 2\}. The original image is resized to 600 pixels in the shorter side. For Stereo RPN, we have 1024 input channels in the final classification and regression layer instead of 512 layers in the implementation \cite{lin2017feature}  due to the concatenation of the left and right feature maps. Similarly, we have 512 input channels in the R-CNN regress head.
		The inference time of Stereo R-CNN for one stereo pair is around 0.28s on the Titan Xp GPU.}
	{\setlength{\parindent}{0cm}
		\subparagraph*{Training.} We define the multi-task loss as:
		\begin{equation}
		\label{eq:loss}
		\renewcommand*{\arraystretch}{1.5}
		\begin{array}{lr}
		{L} = w_{cls}^pL_{cls}^p + w_{reg}^pL_{reg}^p + w_{cls}^rL_{cls}^r + w_{box}^rL_{box}^r \\
		\,\,\,\,\,\,+ \,w_{\alpha}^rL_{\alpha}^r + w_{dim}^rL_{dim}^r + + w_{key}^rL_{key}^r,
		\end{array}
		\end{equation}
		where we use $(\cdot)^p$, $(\cdot)^r$ for representing RPN and R-CNN respectively, and the subscript $box, \alpha, dim, key$ for the loss of stereo boxes, viewpoint, dimension, and keypotin respectively. Each loss is weighted by their uncertainty following \cite{kendall2017multi}. We flip and exchange the left and right image, meanwhile mirror the the viewpoint angle and keypoints respectively to form a new stereo imagery. The origin dataset is thereby doubled with different training targets. During training, we keep 1 stereo pair and 512 sampled RoIs in each mini-batch. We train the network using SGD with a weight decay of 0.0005 and a momentum of 0.9. The learning rate is initially set to 0.001 and reduced by 0.1 for every 5 epochs. We train 20 epochs with 2 days in total.
	}
\begin{figure*}
		\begin{center}
			\includegraphics[width=2.0\columnwidth]{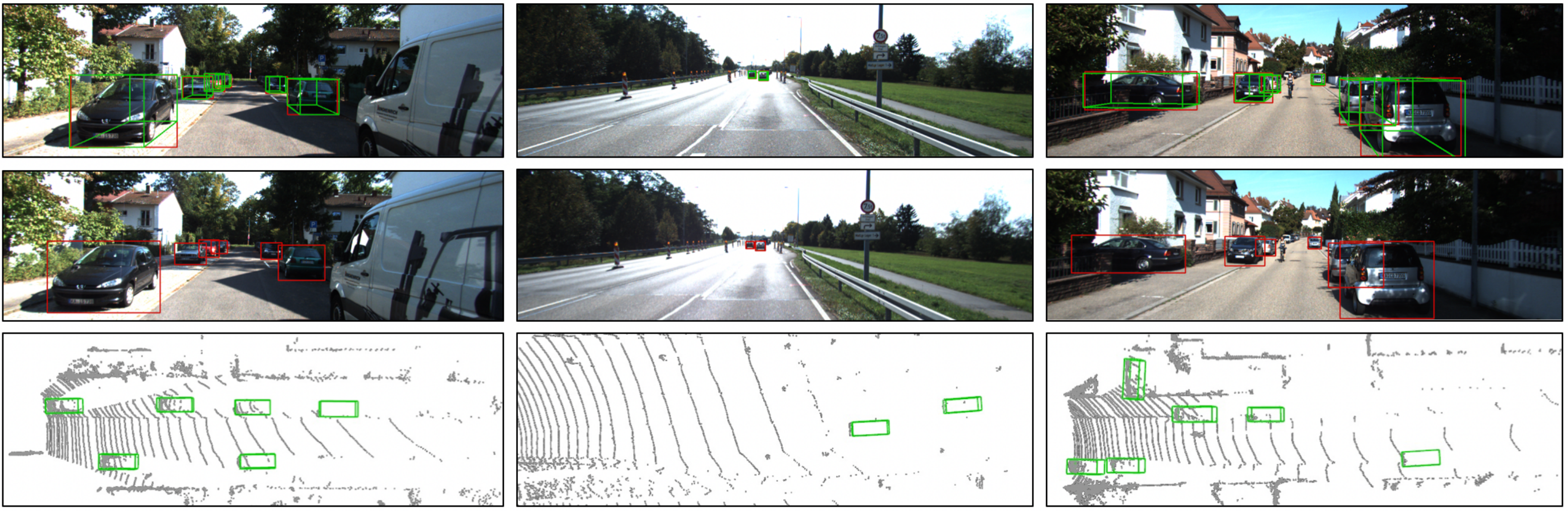}
		\end{center}
		\caption{Qualitative results. From top to bottom: detections on left image, right image, and bird's eye view image.}
		\label{fig:vis}
\end{figure*}
	\section{Experiments}
	We evaluate our method on the challenging KITTI object detection benchmark \cite{geiger2012we}. Following \cite{3dopJournal}, we split 7481 training images into \textit{training set} and \textit{validation set} with roughly the same amount. To fully evaluate the performance of our Stereo R-CNN based approach, we conduct experiments using the 2D stereo recall, 2D detection, stereo association, 3D detection, and 3D localization metrics by comparing with state-of-the-art and self-ablation. Objects are divided into three difficulty regimes: \textit{easy, moderate} and \textit{hard}, according to their 2D box height, occlusion and truncation levels following the KITTI setting.
	
	{\setlength{\parindent}{0cm}
		\subparagraph*{Stereo Recall and Stereo Detection.} Our Stereo R-CNN aims to simultaneously detect and associate object for the left and right image.
Besides evaluating the 2D Average Recall (AR) and 2D Average Precision (AP$_{\rm 2d}$) on both the left and right images, we also define  the stereo AR and stereo AP metrics, where only querying stereo boxes fulfill the following conditions can be considered as the True Positives (TPs):
		\begin{enumerate}
			\itemsep0em
			\item The maximum IoU of the left box with left GT boxes is higher than the given threshold;
			\item The maximum IoU of the right box with right GT boxes is higher than the given threshold;
			\item The selected left and right GT boxes belong to the same object.
		\end{enumerate}
		The stereo AR and stereo AP metrics jointly evaluate the 2D detection and association performance together. As Table.~\ref{table:2d} shows, our Stereo R-CNN has similar proposal recall and detection precision on the single image comparing with Faster R-CNN, while producing high-quality data association in left and right image without additional computation. Although the stereo AR is slightly less than left AR in RPN, we observe almost the same left, right, and stereo APs after R-CNN, which indicates the consistent detection performance on the left and right image and nearly all true positive boxes in the left image have corresponding true-positive right boxes. We also test two strategies for left-right feature fusion: element-wise mean and channel concatenation. As reported in Table.~\ref{table:2d}, the channel concatenation shows better performance since it keeps all the information.
		Accurate stereo detection and association provide sufficient box-level constraints for the 3D box estimation (Sect.~\ref{sec:box}).
	}
\vspace{-0.1cm}
	{\setlength{\parindent}{0cm}
		\subparagraph*{3D Detection and 3D Localization.} We evaluate our 3D detection and 3D localization performance using Average Precision for bird's eye view (AP$_{\rm bv}$) and 3D box (AP$_{\rm 3d}$). Results are shown in Table.~\ref{table:SOTA}, where our method outperforms state-of-the-art monocular-based methods \cite{chen2016monocular,mousavian20173d,cvpr18xu} and stereo-method \cite{3dopJournal} by large margins. Specifically, we outperform 3DOP \cite{3dopJournal} over 30\% for both AP$_{\rm bv}$ and AP$_{\rm 3d}$ across easy and moderate sets. For the hard set, we achieve $\sim$25\% improvements. Although Multi-Fusion \cite{cvpr18xu} obtains significant improvements with stereo input, it still reports much lower AP$_{\rm bv}$ and AP$_{\rm 3d}$ than our geometric method in the moderate set.
Since comparing our approach with LiDAR-based approaches is unfair, we only list one LiDAR-based method VeloFCN \cite{li2016vehicle} for reference, where we outperform it by $\sim$10\% AP$_{\rm bv}$ and AP$_{\rm 3d}$ using IoU = 0.5 in the moderate set.
We also report evaluation results on the KITTI \textit{testing set} in Table.~\ref{tab:test}. The detailed performance can be found online. \footnote{\url{http://www.cvlibs.net/datasets/kitti/eval_object.php?obj_benchmark=3d}}
	}

\begin{figure}
		\begin{center}
			\includegraphics[width=0.95\columnwidth]{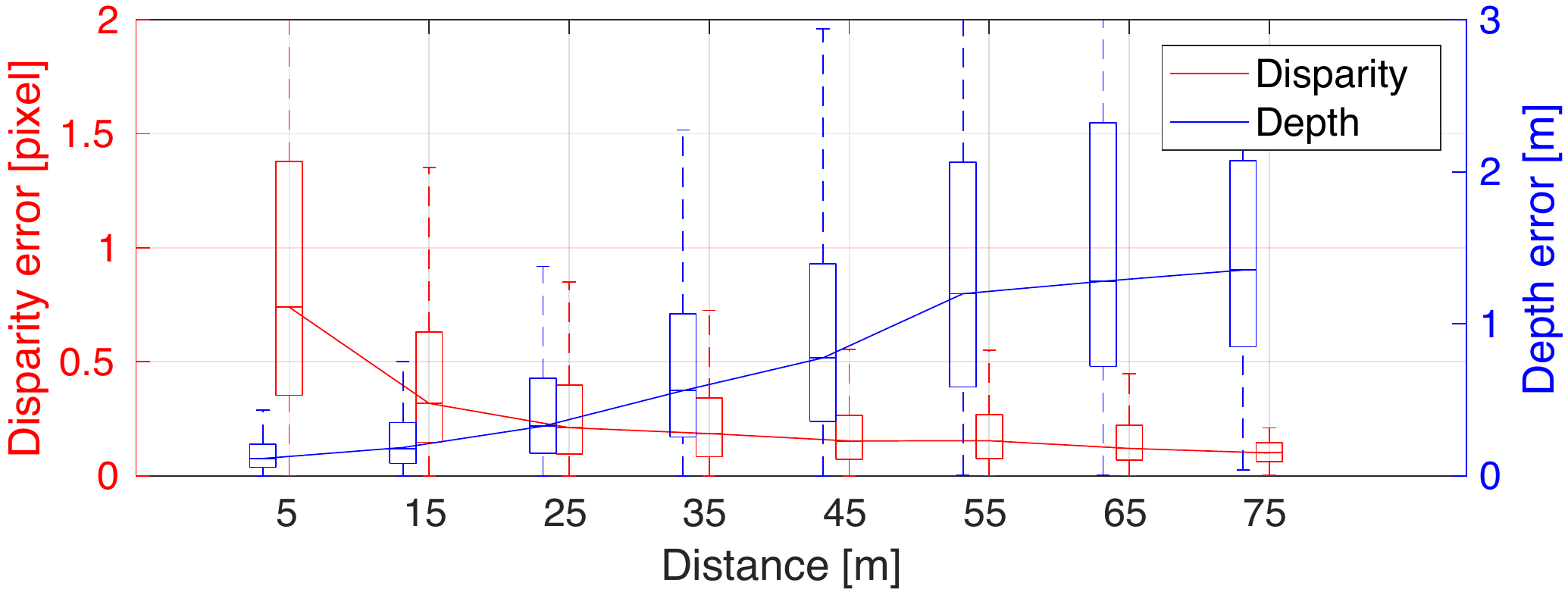}
		\end{center}
		\caption{Relations between the disparity and the depth error with the object distance (best viewed in color). For each distance range ($\pm$5 m), we collect the error statistics for detections with 2D IoU $\ge$ 0.7.}
		\label{fig:depth_dis}
	\end{figure}

	Note that the KITTI 3D detection benchmark is difficult for image-based method, for which the 3D performance tends to decrease as objects distance increases. This phenomenon can be observed  intuitively in Fig.~\ref{fig:depth_dis}, although our method achieves sub-pixel disparity estimation (less than 0.5 pixel), the depth error becomes larger as the object distance increase due to the inversely proportional relation between disparity and depth. For objects with explicit disparity, we achieve high accurate depth estimation based on rigorous geometric constraints. That explains why a higher IoU threshold, an easier regime the object belongs, we obtain more improvements compared with other methods.
	\begin{table}
	\setlength{\belowcaptionskip}{-0.3cm}
		\begin{center}
			\renewcommand{\arraystretch}{1.3}
			\resizebox{0.41\textwidth}{!}{%
				\begin{tabular}{l|ccc|ccc}
					\, & \multicolumn{3}{c|}{$\rm AP_{bv}$ (IoU=0.7)} & \multicolumn{3}{c}{$\rm AP_{3d}$ (IoU=0.7)}\\
					\cline{2-7}
					Method & \,Easy\, & Mode & \,Hard\, &  \,Easy\, & Mode & \,Hard\, \\
					\Xhline{1pt} 
					Ours  & 61.67 & 43.87 & 36.44 & 49.23 & 34.05 & 28.39 \, \\
			\end{tabular}}		
		\end{center}
		\caption{3D detection and localization AP on the KITTI \textit{test set}.}
		\label{tab:test}
		\vspace{-0.3cm}
	\end{table}	
	\begin{table*}
		\begin{center}
			\renewcommand{\arraystretch}{1.2}
			\resizebox{0.85\textwidth}{!}{%
				\begin{tabular}{lc|c|ccc|ccc|ccc|ccc}
					\, & \, & \, &
					\multicolumn{3}{c}{AP$\rm _{bv}$ (IoU=0.5)} & \multicolumn{3}{c|}{AP$\rm _{bv}$ (IoU=0.7)} & \multicolumn{3}{c}{AP$\rm _{3d}$ (IoU=0.5)} & \multicolumn{3}{c}{AP$\rm _{3d}$ (IoU=0.7)} \\
					\cline{4-15}
					Flip & Uncert & AP$\rm _{2d}^{0.7}$ & Easy\, & Mode & \,Hard\, &  \,Easy\, & Mode & \,Hard\, & \,Easy\, & Mode & \,Hard\, &  \,Easy\, & Mode & \,Hard \, \\
					\Xhline{1pt} 
					 \, & \, &  79.03 & 76.82 & 64.75 & 54.72 & 54.38 & 36.45 & 29.74 & 75.05 & 60.83 & 47.69 & 32.30 & 21.52 & 17.61\, \\
					\hline
					  \checkmark & \, &  79.78 & 78.24 & 65.94 & 56.01 & 60.93 & 40.33 & 33.89 & 76.87 & 61.45 & 48.18 & 40.22 & 28.74 & 23.96\, \\
					 \, & \checkmark &  88.52 & 84.89 & 67.02 & 57.57 & 60.93 & 40.91 & 34.48 & 78.76 & 64.99 & 55.72 & 47.53 & 30.36 & 25.25\, \\
					 \checkmark & \checkmark & \textbf{88.82} & \textbf{87.13} & \textbf{74.11} & \textbf{58.93} & \textbf{68.50} & \textbf{48.30} & \textbf{41.47} & \textbf{85.84} & \textbf{66.28} & \textbf{57.24} & \textbf{54.11} & \textbf{36.69} & \textbf{31.07}\, \\
				\end{tabular}
			}		
		\end{center}
		\caption{Ablation study of using flip augmentations and uncertainty weight, evaluated on KITTI \textit{validation set}.}
		\label{table:ablation}
	\end{table*}
\vspace{-0.2cm}
	{\setlength{\parindent}{0cm}
		\subparagraph*{Benefits of the Keypoint.} We utilize the 3D box estimator (Sect.~\ref{sec:box}) to calculate the coarse 3D box and rectify the actual 3D box after the dense alignment. An accurate 3D box estimator is thereby important for the final 3D detection. To study benefits of the keypoint for 3D box estimator, we evaluate the 3D detection and 3D localization performance without using the keypoint, where we use the regressed viewpoint to determine the relations between 3D box corners and 2D box edges, and employ Eq.~\ref{eq:alpha} to constraint the 3D orientation for all objects. As reported in Table.~\ref{tab:orien}, the usage of the keypoint improve both AP$_{\rm bv}$ and AP$_{\rm 3D}$ across all difficulty regimes by non-trivial margins. As the keypoint provides pixel-level constraints to the 3D box corner in addition to the 2D box-level measurements, it ensures more accurate localization performance.
	}
	\begin{table}
		\begin{center}
			\renewcommand{\arraystretch}{1.3}
			\resizebox{0.47\textwidth}{!}{%
				\begin{tabular}{l|ccc|ccc}
					\, & \multicolumn{3}{c|}{w/o Keypoint} & \multicolumn{3}{c}{w/ Keypoint}\\
					\cline{2-7}
					Metric & \,Easy\, & Mode & \,Hard\, &  \,Easy\, & Mode & \,Hard\, \\
					\Xhline{1pt} 
					$\rm AP_{bv}$ (IoU=0.5)  & 87.10 & 67.42 & 58.41 & \textbf{87.13} & \textbf{74.11} & \textbf{58.93} \, \\
					$\rm AP_{bv}$ (IoU=0.7)  & 59.45 & 40.44 & 34.14 & \textbf{68.50} & \textbf{48.30} & \textbf{41.47} \, \\
					\hline
					$\rm AP_{3d}$ (IoU=0.5)  & 85.21 & 65.23 & 55.75 & \textbf{85.84} & \textbf{66.28} & \textbf{57.24} \, \\
					$\rm AP_{3d}$ (IoU=0.7)  & 46.58 & 30.29 & 25.07 &  \textbf{54.11} & \textbf{36.69} & \textbf{31.07} \, \\
			\end{tabular}}		
		\end{center}
		\caption{Comparing 3D detection and localization AP of w/o and w/ keypoint, evaluated on KITTI \textit{validation set}.}
		\label{tab:orien}
	\end{table}
	\begin{table}
		\begin{center}
			\renewcommand{\arraystretch}{1.3}
			\resizebox{0.41\textwidth}{!}{%
				\begin{tabular}{l|c|cc|cc}
					Config & Set & {AP$_{\rm bv}^{0.5}$} &{AP$_{\rm bv}^{0.7}$} & {AP$_{\rm 3d}^{0.5}$} & {AP$_{\rm 3d}^{0.7}$}\\
					\Xhline{1pt} 
					\multirow{3}{*}{w/o Alignment} & Easy & 45.59 & 16.87 & 41.88 & 11.37 \\
					\, &  Mode & 33.82 & 10.40 & 27.99 & 7.75 \\
					\, &  Hard & 28.96 & 10.03 & 22.80 & 5.74 \\
					\hline
					\multirow{3}{*}{\shortstack{w/ Alignment\\w/o 3D rectify}} & Easy & 86.15 & 66.93 & 83.05 & 48.95 \\
					\, &  Mode & 73.54 & 47.35 & 65.45 & 32.00 \\
					\, &  Hard & 58.66 & 36.29 & 56.50 & 30.12 \\
					\hline
					\multirow{3}{*}{\shortstack{w/ Alignment\\w/ 3D rectify}} & Easy & \textbf{87.13} & \textbf{68.50} & \textbf{85.84} & \textbf{54.11} \\
					\, &  Mode & \textbf{74.11} & \textbf{48.30} & \textbf{66.28} & \textbf{36.69} \\
					\, &  Hard & \textbf{58.93} & \textbf{41.47} & \textbf{57.24} & \textbf{31.07} \\
					
			\end{tabular}}		
		\end{center}
		\caption{Improvements of using our dense alignment and 3D box rectify, evaluated on KITTI \textit{validation set}.}
		\label{table:alignment}
	\end{table}
	\vspace{-0.2cm}
	{\setlength{\parindent}{0cm}
		\subparagraph*{Benefits of the Dense Alignment.} This experiment shows how significant improvements the dense alignment brings. We evaluate the 3D performance of the coarse 3D box (w/o Alignment), for which the depth information is calculated from box-level disparity and 2D box size. Even if 1-pixel disparity or 2D box error will cause large distance error for distant objects. In result, although the coarse 3D box has a precise projection on the image as we expected, it is not accurate enough for 3D localization. Detailed statistics can be found in Table.~\ref{table:alignment}. After we recover the object depth using the dense alignment and simply scaling the $x,y$ (w/ Alignment, w/o 3D rectify), we obtain major improvements on all the metric.
		Furthermore, when we using the box estimator (Sect.~\ref{sec:box}) to rectify the entire 3D box by fixing the aligned depth, the 3D localization and 3D detection performance are further improved by several points.
	}
	\vspace{-0.2cm}
	{\setlength{\parindent}{0cm}
		\subparagraph*{Ablation Study.} We employ two strategies to enhance our model performance. To validate the contributions of each strategy, we conduct experiments with different combinations and evaluate the detection and localization performance. As Table.~\ref{table:ablation} shows, we use Flip and Uncert to represent the proposed stereo flip augmentation and the uncertainty weight for multiple losses \cite{kendall2017multi}. Without bells and whistles, we already outperform all state-of-the-art image-based methods. Each strategy further enhances our network performance by several points. Detailed contributions can be found in Table.~\ref{table:ablation}.
Balancing the multi-task loss using uncertainty weight yields non-trivial improvements in both 3D detection and localization tasks. With stereo flip augmentation, the left-right images are flipped and exchanged, and the training target for the the perspective keypoint and viewpoint are also changed respectively. Therefore the training set is doubled with different inputs and training targets.
 Combining two strategies together, our method obtains strongly promising performance in both 3D detection and 3D localization tasks (Table.~\ref{table:SOTA}).
	}
		{\setlength{\parindent}{0cm}
		\subparagraph*{Qualitative Results.} We show some qualitative results in Fig.~\ref{fig:vis}, where we visualize corresponding stereo boxes on the left and right images. The 3D box is projected to the left and bird's eye view image respectively. Our joint sparse and dense constraints ensure the detected box is well aligned on both image and LiDAR point cloud.
	}
	\section{Conclusion and Future Work}
	In this paper, we propose a Stereo R-CNN based 3D object detection method in autonomous driving scenarios. Formulating the 3D object localization as a learning-aided geometry problem, our approach takes the advantage of both semantic properties and dense constraints of objects. Without 3D supervision, we outperform all existing image-based methods by large margins on 3D detection and 3D localization tasks, and even better than a baseline LiDAR method \cite{li2016vehicle}. 
	
	Our 3D object detection framework is flexible and practical where each module can be extended and further improved. For example, Stereo R-CNN can be extended for multiple object detection and tracking. We can replace the boundary keypoints with instance segmentation to provide more precise valid RoI selection. By learning the object shape, our 3D detection method can be further applied to general objects.

{\setlength{\parindent}{0cm}		
		\subparagraph*{Acknowledgment.} This work was supported by the Hong Kong Research Grants Council Early Career Scheme under project 26201616.		
		}

	{\small
		\bibliographystyle{ieee}
		\bibliography{egbib}
	}
	
\end{document}